\documentclass[runningheads]{llncs}
\usepackage[T1]{fontenc}
\usepackage[normalem]{ulem}
\usepackage{booktabs}
\usepackage{xcolor}
\usepackage{graphicx}
\begin{document}
\title{Visual graphs for image classification: does the structure affect performance?}
%
%\titlerunning{Abbreviated paper title}
% If the paper title is too long for the running head, you can set
% an abbreviated paper title here
%
\author{Alessandra Ibba\inst{1}}
\authorrunning{A. Ibba}
% First names are abbreviated in the running head.
% If there are more than two authors, 'et al.' is used.
%
\institute{University of Sassari - Department of Engineering, Sassari, Italy 
\email{\{a.ibba26\}@studenti.uniss.it}} 
\maketitle              % typeset the header of the contribution
\begin{abstract}

Deep learning models have emerged in machine learning and related fields, demonstrating astonishing performance in various visual tasks. Despite their great success, however, these models are unable to fully encode intrinsic visual structures, and often ignore the spatial, topological, and semantic information contained within an image. Graph neural networks offer a good framework to face this aspect, but their effective use for visual tasks has been only partly explored and mainly starting from a limited perspective. 
This work aims to address this gap by conducting a systematic comparison
of current graph construction techniques within the context of a fixed three-layer GCN architecture.
Through an empirical study, it demonstrates in particular how the network structure
affects performance and provides an important methodological contribution
regarding the computational stages preceding graph utilization, which will be strongly influenced by the structure itself.
\end{abstract}
\section{Introduction}
In recent years, both computer vision and natural language processing have greatly benefited from the success of convolutional neural networks (CNNs) \cite{zhao2024review}. A key factor contributing to this success is the ability of CNNs to exploit the inherent structural properties of specific data types. In the case of images, their grid-like organization allows convolutional layers to extract high-level features and build hierarchical representations, thereby providing substantial expressive power. Despite these advantages, CNNs often struggle to fully capture the relationships among objects within a scene and to effectively incorporate contextual information, which can be crucial in many applications. Moreover, CNN-based solutions are typically characterized by a large number of parameters and high computational complexity, regardless of the actual image content. Vision Transformers (ViTs) \cite{dosovitskiy2021imageworth16x16words} partially address these issues by implementing self-attention mechanisms to directly model the relationships between pairs of image patches. However, they fail to fully preserve spatial layout; furthermore, they are unable to handle fine details and precise localization due to their very nature, which relies on patches of significant size \cite{Han2022VisionGNN}. In addition, like CNNs, ViTs are parameter-heavy and computationally demanding.
Graph neural networks offer a compelling framework to address these limitations. By operating on non-regular graph structures, they are able to model structural relationships within the data and reveal intrinsic connections among different parts of an image. However, the irregular nature of graphs also poses challenges, as convolution and filtering operations on graphs are not as straightforward or well-defined as they are in the image domain \cite{Zhang2019}. Graph Convolutional Networks (GCN), introduced by \cite{kipf2017semisupervisedclassificationgraphconvolutional}, are an efficient variant of convolutional neural networks for graphs. The input data of a GNN is a graph, that is a graph-structured representation of visual data in which nodes correspond to visual entities and edges encode relationships between them, such as spatial adjacency, geometric proximity, semantic similarity, or contextual interaction. Note that the set of nodes typically carries visual features extracted from images but these features can vary greatly from simple pixels to regions or other structured representations (sometimes denoted as descriptors) of image areas. Information carried by nodes and edges commonly takes the name of ``embedding''; from an intuitive perspective, a node embedding answers to the basic question ``what is a visual element'' while an edge embedding determines ``how two visual elements are related''.

Three are the main challenges in applying GCNs to computer vision. First, defining an appropriate graph representation from an image is not straightforward, as it requires specifying nodes and edges in a way that can significantly influence all subsequent processing stages. Second, it is necessary to select meaningful embeddings to be associated with the graph elements and structure in order to effectively address specific computer vision tasks. Third, the practical use of graph-based models raises a number of concerns related to  computational efficiency, scalability and data requirements so as to stability with respect to noise and dynamic visual scenes.

This work focuses on the first two problems listed above: the graph construction and the mechanisms by which information is associated with the graph structure. In particular, it investigates how graph topology and structural complexity affect the performance of GNNs, and compares different types of node and edge information, showing that these are less influential from an overall performance perspective.
The following sections provide an overview of the state of the art; the methodology adopted for the comparison is then described, followed by a discussion of the results obtained.

\section{State of the art}
The methodologies to define nodes and edges of an image can be broadly categorized into classical computer vision techniques and more recent deep learning-based approaches.
\subsection{Nodes definition}
 Classic nodes definition techniques are essentially based on a wide range of well-known and popular solutions in computer vision and image processing, including: 
\begin{enumerate}
    \item \underline{pixel nodes}: this is the most straightforward approach which treats each pixel as a node and often results in computationally prohibitive graphs \cite{edwards2016graphbasedconvolutionalneural}. As an alternative, the final number of vertices can be established in advance and linked to a regular structure. The pixels are then assigned to the nearest vertex, contributing to define its characteristics. At the end of the process, the original image is segmented into regular sub-regions, each of which corresponds to a single vertex of the network \cite{NEURIPS2018_4efb80f6} \cite{Han2022VisionGNN}. 
    \item \underline{superpixel nodes}: in this case sets of similar pixels are grouped by a segmentation algorithm defining different superpixels. Each superpixel then becomes a node. Images can be segmented using different thresholds and different grouping methods, resulting in a varying number of nodes and pixel density per node. The structure of the graph is necessarily non regular. Additional pieces of information, such as color tones, luminance and texture, can be used to represent the characteristics of superpixels.  Examples of superpixel extraction algorithms are SLIC, Quickshift and Felzenszwalb \cite{nazir2021surveyimagebasedgraph}. SLIC, as used by \cite{Rodrigues2024GraphCN}, takes the desired number of superpixels as input. In this context, compactness is calculated as a ratio of the perimeter to the area. Thus, SLIC adapts this parameter based on the region’s texture, yielding regularly shaped superpixels regardless of texture.
    \item \underline{interest points}: key points or salient points are distinct, repeatable, and unique points within an image where the signal shows a significant and recognizable variation. They are often associated with significant luminance transitions such as corners or spots, but there are many different technical approaches in the literature that seek to improve the repeatability of detection even under different imaging conditions. Furthermore, robust point of interest detectors must handle digital artifacts, noise, poor resolution and optical distortions \cite{jing2022imagefeatureinformationextraction}. 
    Techniques like the graph-based visual saliency model \cite{NIPS2006_4db0f8b0} generate a saliency map by combining responses across multiple feature channels (intensity, color, orientation); the most informative points are designated as candidate nodes, thereby bringing with them an aggregate of information dependent on the feature channels under consideration.
\end{enumerate}

\subsection{Edges definition} 
Once nodes are defined, the next critical step is to establish edges that capture the relational structure between them. Common methodologies include K-NN and RAG, as described below:

\begin{enumerate}
    \item \underline{K-nearest neighbors (K-NN)}: creates edges by connecting each node to its K most similar neighbors. The definition of "similarity" is crucial and can be based on single or combined metric. A single metric uses a sole distance measure; not only geometric centroids, to define euclidean distances between nodes also other kinds of metrics have been explored considering just luminance, color or other local image features \cite{NEURIPS2018_4efb80f6} \cite{knyazev2019imageclassificationhierarchicalmultigraph} \cite{10.1007/s00521-022-07368-1}. Combined metric integrates multiple factors such as spatial and color distances. Euclidean distance or cosine similarity can be used to compare feature values, taking into account both structural and semantic relationships \cite{10.5555/3018843.3018851}. 
  
    \item \underline{Region adjacency generation (RAG)}: mainly used in graphs built from superpixels, uses edges to represent spatial adjacency between regions. Adjacency is determined by shared boundaries between neighboring regions in the image \cite{9621101} \cite{nazir2021surveyimagebasedgraph}.
\end{enumerate}
\subsection{Deep neural networks graph definition}
Approaches based on deep neural networks have more recently demonstrated an impressive ability to extract graphs from images. They operate defining nodes, edges and embeddings simultaneously. In this respect two are the main options considered in the literature:
\begin{enumerate}
    \item \underline{CNN}: feature maps extracted from pre-trained CNNs are used to define and characterize nodes. More specifically, each node of the graph is associated to a feature embedding while the GCN directly maps these embeddings into a set of inter-dependent classifiers, which can be directly applied to image features for classification \cite{chen2019multilabelimagerecognitiongraph}.  
    \item \underline{ViT}: the image is decomposed into a set of patches, representing  regions of interest. Each region becomes a graph node. To this end, mechanisms like Spatial Transformer (ST) or Semantic Attention Module (SAM) are applied, with a fixed-output size defined at the beginning and representing the number of desired nodes  \cite{zhao2021transformerbaseddualrelationgraph}. 
\end{enumerate} 
\subsection{Consolidated approaches in full graph construction}
The construction of the graph involves a combination of the techniques described above, with the ultimate aim of defining the overall structure and the corresponding embeddings.

Han and colleagues \cite{Han2022VisionGNN} divide the image into N regularly shaped patches and associate to each patch a feature vector consisting on the image patch itself. These features are viewed as a set of unordered nodes. For each node, k-nearest neighbors are find and a directed edge from each node to all the neighbor nodes added. 
Ma and colleagues \cite{ma2023imagesetpoints} propose a method to convert the image into a collection of points, each point containing both color and position information. The point cloud is down-sampled for efficiency; then, with a context cluster operation, data points are grouped by selecting a number of "anchor" points evenly in space and, for each anchor, finding its k nearest neighboring points. The k points are fused into one new point using a linear projection. Feature points are grouped into clusters based on cosine similarity, which measures the angle between feature vectors. Crucially, because each point's feature vector is influenced by its coordinates, this similarity measure implicitly considers both visual similarity (color/texture) and spatial proximity. 
The approach suggested by Campos et al. \cite{10.1007/978-3-030-87897-9_2} exploits features extracted through pre-trained convolutional neural networks. First, a GNN is defined for each image with fully connected nodes. Then, the GNN is tuned by pruning the edges, considering the dissimilarity between nodes. Finally, experiments are performed considering different types of deep features aggregated with the GNN, showing that tuned GNNs, with a significantly decreased number of edges, can achieve the same level of accuracy as GNNs with complete and random connections.
In order to leverage useful information from multiple relationships, Knyazev and colleagues \cite{knyazev2019imageclassificationhierarchicalmultigraph} compute superpixels at several resolution scales and define spatial and hierarchical edges. The experiments carried on three different datasets (MNIST, CIFAR-10 and PASCAL) show  that intensity values of superpixels and their coordinates are rather weak features that  carry only limited geometry and texture information. 
Linh and Youn \cite{9621101} also exploit superpixels  by proposing a dynamic approach that generates a new set of edges based on the current representations of the points using the K-nearest neighbour sampling method. Although their model does not outperform traditional CNN models, it demonstrates that  it is possible to achieve good accuracy while reducing the complexity of the neural model.

\subsection{Graph sparsification}
Although raw graphs can be mapped directly to a network, their inherent complexity often hinders their practical application. To address this issue,  sparsification techniques can be applied that streamline the graph by removing redundant elements, all the while strictly maintaining essential topology and connectivity.
Node-centric pruning  reduces the size and complexity of a network by identifying and removing  nodes that are  not critical to the graph’s structure \cite{Shokouhinejad24}; edge-centric pruning focus instead on the elimination of non-essential connections. These are often based on thresholding or topological criteria. The aim is usually to create a sparser graph that is more computationally efficient while retaining the structural backbone of the data.
Kosman and collegues \cite{kosman2021lspaccelerationregularization} show that graph sparsification can greatly improve the performance of GNNs in tasks where the graph topology is significant. To this purpose they introduce Locality Sensitive Pruning, a new algorithm for edges pruning based on locality sensitive hashing.

The next section describes the experimental architecture, specifically designed to enable the systematic comparison between various image-to-graph construction and feature-extraction pipelines, with explicit reference to the image classification problem. The work focuses on the structural complexity of the network and the type of information associated with nodes and edges. Note that, as the primary objective of the study is to analyse networks that differ structurally yet are comparable in terms of number of nodes, edge embeddings are disregarded and the GNN is trained to generate meaningful representations regardless of the initial edge weights, which are consequently randomized.
\begin{figure}[ht!]
\centering
{\includegraphics[width=0.80\linewidth]{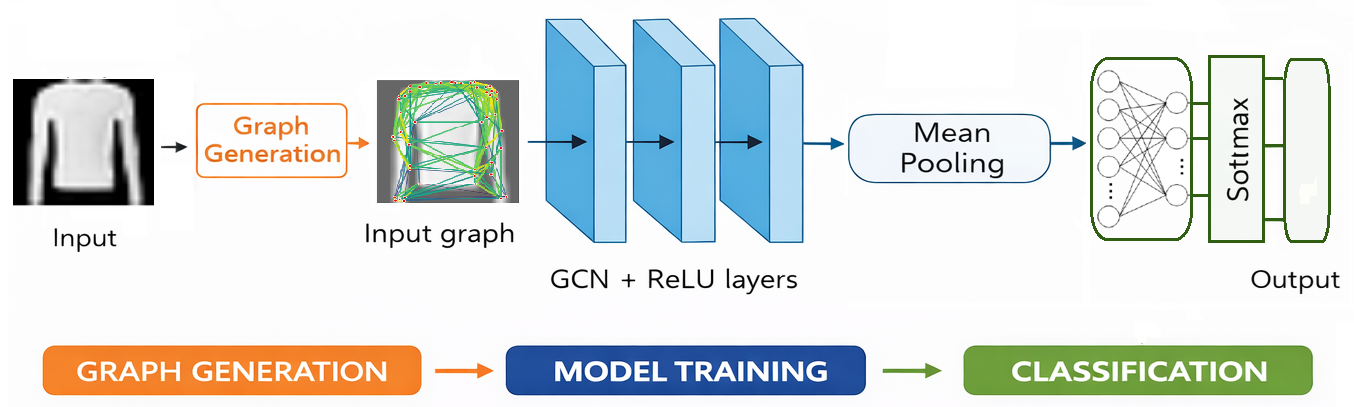}}
\caption{Architecture of the model}
\label{fig:model}
\end{figure}
\section{Method}
\subsection{The model}
 The experimental architecture in the comparative model consists of a graph generation layer, followed by three sequential GCN+ReLU Activation layers. This sequence of layers is followed by a global mean pooling stage which in turn feeds a softmax layer. The main components of the architecture are shown in Fig. \ref{fig:model}

As highlighted by Rodriguez and Carbonera \cite{Rodrigues2024GraphCN}, the performance of a GNN on a visual task is profoundly influenced by the initial graph generation stage. Following a superpixel approach, they propose to evaluate the impact of the graph building choices using: the degree of segmentation (number of nodes in the resulting graph), the selection of features for characterizing each node and the method for defining edges. They also show that the performance gain saturates increasing the number of nodes over the limit of 10\%  of the total number of image pixels. 

In our experimental model the maximum number of nodes is fixed to 50 (N = 50), corresponding to about 6.4\% of the total number of image pixels. We thus focus  on  three parametric dimensions: spatial position of the nodes, image features associated to each node (embeddings) and graph construction and sparsification.

Spatial position is computed considering three different node extractors: a regular grid with 7x7 nodes (as regular structure coming from deep neural networks analyzers), superpixel (SLIC method) and interest point (Harris method from Kornia library). 

For each node, a feature vector is defined  by concatenating visual descriptors with their normalized spatial coordinates. Three descriptors are considered: ViT-based, CNN-based and  gradient‑based (SIFT‑like), resulting in dimensions of 768, 256, and 130 floats, respectively. Specifically, the ViT features include 766 visual features extracted from a pretrained ``ViT tiny patch16 224" model \cite{wu2022tinyvit} while the CNN information comprise 254 visual features, from a pretrained ResNet18 model \cite{he2016deep}. Gradient‑based features are based on the well known work of Lowe \cite{Lowe2004} and include 16 gradient orientation histograms along eight directions, for a total number of 128 visual features.

Concerning edges definition the "base" structure is obtained by computing, for each node, the six nearest neighbors (K-NN with K=6). Note that the distance is directly computed on node embeddings thus considering  both spatial and visual information.
Starting from the base structure, three levels of pruning are applied; the first level simply consider the average distance between all the nodes of the graph, pruning all edges above this value. The effect of this approach is that the graph is moderately simplified: isolated nodes may appear and  therefore must be removed. %At a second level, pruning follows a shortest path approach removing all the edges that  do not contribute to an efficient (low distance) connection.In this case all the nodes are retained but the number of edges  drops depending on the image information. 
At a second level, in order to compare sparsification to most recent advances, locality sensitive pruning is considered \cite{kosman2021lspaccelerationregularization}. Finally, a global sparsification step is applied by extracting a minimum spanning tree (MST) from the original graph using Kruskal’s algorithm. Edges are sorted according to their weights and iteratively selected while avoiding cycles. This procedure preserves only the strongest connections present in the initial graph, ensuring minimal connectivity while maintaining the original edges.
Table \ref{tab:dimensions} better summarizes the notation and the meaning of the parametric dimensions explored in the following.
\begin{table}
  \caption{Parametric dimensions}
  \label{tab:dimensions}
  \centering
  \begin{tabular}{llcc}
    \toprule
     Spatial position & Description\\
    \midrule
    GRID & Grid-based (7x7 nodes)\\
    SP & Superpixel SLIC extraction\\
    IP & 50 most salient points\\   
     \toprule
     Feature type & Description\\
    \midrule
    ViT & 768-dimensional  vector \\
    CNN & 256-dimensional  vector \\
    SIFT & 130-dimensional vector\\   
     \toprule
     Graph model & Description\\
    \midrule
    BM & Base model (6-NN)\\
    LDM & Low Distance over graph mean\\
    LSP & Local sensitive pruning\\
    MinCONN & Minimal connectivity on original edges\\    
    \bottomrule
  \end{tabular}
\end{table} 
% Please note that each graph under test is designated by a short acronym and that the characteristics of the resulting structure are summarised by the following indicators  \cite{KAVIANI2021115073}:
%  \begin{enumerate}
%     \item Number of nodes N
%     \item Number of edges E
%     \item Density D
%     \item Average path length L
%     \item Clustering coefficient C   
% \end{enumerate}
% Also note that the measures listed above reveal not only the basic structure of the graph (N, E and D), but also the presence of edges connecting spatially distant nodes (a low value of L is a characteristic of ‘small-world’ networks) and modularity (a high value of C indicates the presence of cliques, as opposed to random connections). These aspects will be better commented in the next section.
\section{Experiments}
To evaluate the impact of different graph generation strategies, we conducted experiments on the fashion-MNIST dataset \cite{xiao2017fashionmnistnovelimagedataset}. The dataset comprises 70,000, 28x28 pixels, grayscale images of fashion products from 10 categories.
The graphs, generated by varying the parameters previously described, were used to train the GCN and classification layers. The Adam optimizer was used for training, with a fixed learning rate of 0.001. We adopted a stratified 10-fold cross-validation procedure on the training set only. In each fold, nine‑tenths of the training data are used for training and one‑tenth for validation, preserving class distributions. Training runned for up to 180 epochs with early stopping (patience 15) based on validation accuracy. After cross‑validation, the model was retrained on the full training set for the average number of epochs that yielded the best validation accuracy across folds, and finally evaluated on a separate held‑out test set.

\subsection{Analysis of graph construction results} 
\begin{table}
\caption{Base Model: graph structure indicators derived from the construction phase (average values over 10 classes of the fashionMNIST dataset). E represents the total number of connections established in the graph after applying a mutual k-nearest neighbors strategy. Network density D is the ratio between the graph edges to the maximum possible edges. Average path length L is the mean distance between two nodes, averaged over the distances between all nodes in the graph. Clustering coefficient C is the probability that two neighbors of a node are connected together.}
\label{tab:BaseModel}
\setlength{\tabcolsep}{8pt}
\centering
 \begin{tabular}{ll|ccccc}
\toprule
\textbf{Feature} & \textbf{Node method} & \textbf{N} & \textbf{E} & \textbf{D} & \textbf{L} & \textbf{C} \\
\midrule
ViT   & GRID           & 49 & 588 & 0.250 & 2.987 & 0.550 \\
      & SUPERPIXEL     & 40 & 462 & 0.323 & 2.551 & 0.589 \\
      & INTEREST POINT & 50 & 597 & 0.247 & 2.698 & 0.567 \\
\midrule
CNN   & GRID           & 49 & 588 & 0.250 & 2.202 & 0.646 \\
      & SUPERPIXEL     & 40 & 462 & 0.323 & 2.243 & 0.580 \\
      & INTEREST POINT & 50 & 597 & 0.247 & 2.762 & 0.595 \\
\midrule
SIFT  & GRID           & 49 & 588 & 0.250 & 2.457 & 0.641 \\
      & SUPERPIXEL     & 40 & 462 & 0.323 & 2.207 & 0.654 \\
      & INTEREST POINT & 50 & 597 & 0.247 & 2.440 & 0.467 \\
\bottomrule
\end{tabular}
\end{table}

\begin{figure}[ht!]
\centering
{\includegraphics[width=0.7\linewidth]{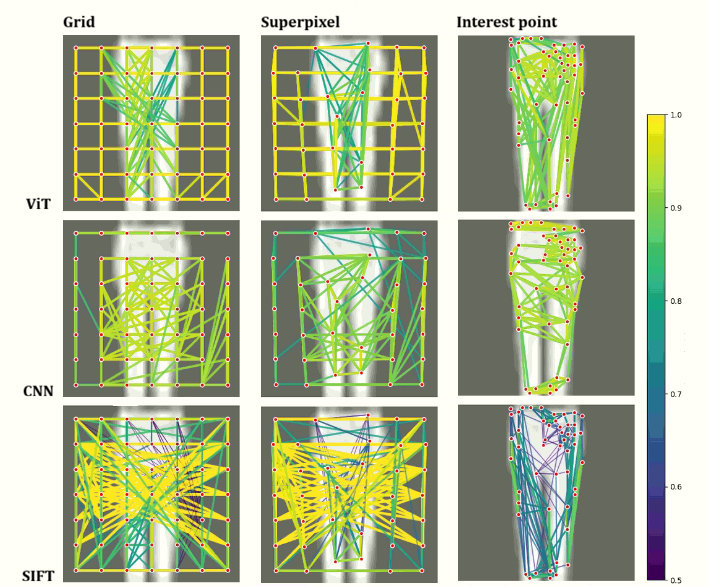}}
\caption{Base Model: final structure of the graph derived from the construction phase.}
\label{fig:BaseModel}
\end{figure}
 Fig. \ref{fig:BaseModel} shows the final base structure of the graph for one of the classes of the fashion-MNIST database (trouser). The three node extraction methods exhibit fundamentally various spatial distributions, as expected. The uniform and rigid spatial coverage emerges with Grid node extractor. The Interest point extractor shows markedly different behavior, with concentrated nodes in salient regions. While superpixel extractor offers a balance, between the uniformity of Grid sampling and the concentration of Interest points, node positioning. Table \ref{tab:BaseModel} summarizes the results of the graph building phase for the Base Model. 
 
\begin{figure}[ht!]
\centering
{\includegraphics[width=0.7\linewidth]{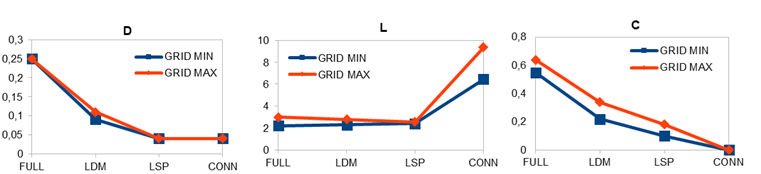}}
\caption{Line charts with markers of D, L and C parameters for node extractor GRID.}
\label{fig:Andamenti}
\end{figure}

 Fig. \ref{fig:Andamenti} shows the behavior of D, L and C for the base graph from the node extractor 'grid' and for the filtered graphs. The behavior is similar for the graphs obtained using the ``superpixel'' and ``interest point'' node extractors. Note how filtering acts on the decrease of D and C while in the case of L it causes a slight decrease in the case of LDM and LSP graphs and a notable increase in the case of MinCONN graphs. The combination of a high clustering C and a low path length L observed in the Base Model graphs suggests the presence of small-world properties. With significantly lower clustering coefficients and longer path lengths, the LDM graphs are structurally similar to sparse random networks or dilated regular structures. As evidenced by a clustering coefficient of zero, the MinCONN graphs are acyclic (i.e., trees). These structures ensure global connectivity with a minimal number of edges, resulting in the longest characteristic path lengths observed in the study. This behavior is consistent with the sparsification performed.

\subsection{Performance analysis}
\begin{table}
  \caption{Classification performance - Nmax = 50, E = 6-nneigh}
  \label{tab:Fashion MNIST Results}
  \setlength{\tabcolsep}{6pt}
  \centering
  \begin{tabular}{lll|cc}
    \toprule
    Node Method & Features & TopPerforming & Acc & F1\\
    \specialrule{1.2pt}{0.5ex}{0.5ex}
    SUPERPIXELS & CNN & MinCONN & 0.8364 & 0.8318\\
    &  & base & 0.8361 & 0.8353\\
    \midrule
    & VIT & MinCONN & 0.8572 & 0.8581\\
    &  & LSP & 0.8543 & 0.8558\\
    \midrule
    & SIFT & MinCONN & 0.8089 & 0.8072\\
    &  & base & 0.7929 & 0.7912\\
    \specialrule{1.2pt}{0.5ex}{0.5ex}
    INTEREST POINTS & CNN & MinCONN & 0.8511 & 0.8498\\
    &  & LDM & 0.8450 & 0.8438\\
    \midrule
    & VIT & MinCONN & 0.8681 & 0.8672\\
    &  & base & 0.8623 & 0.8616\\
    \midrule
    & SIFT & LDM & 0.7659 & 0.7611\\
    &  & base & 0.7624 & 0.7497\\
    \specialrule{1.2pt}{0.5ex}{0.5ex}
    GRID & CNN & LSP & 0.8537 & 0.8529\\
    &  & base & 0.8424 & 0.8406\\
    \midrule
    & VIT & MinCONN & 0.8559 & 0.8558\\
    &  & base & 0.8510 & 0.8533\\
    \midrule
    & SIFT & base & 0.8284 & 0.8247\\  
    &  & LDM & 0.8165 & 0.8130\\
    \bottomrule
  \end{tabular}
\end{table}
\begin{figure}[ht!]
\centering
{\includegraphics[width=0.98\linewidth]{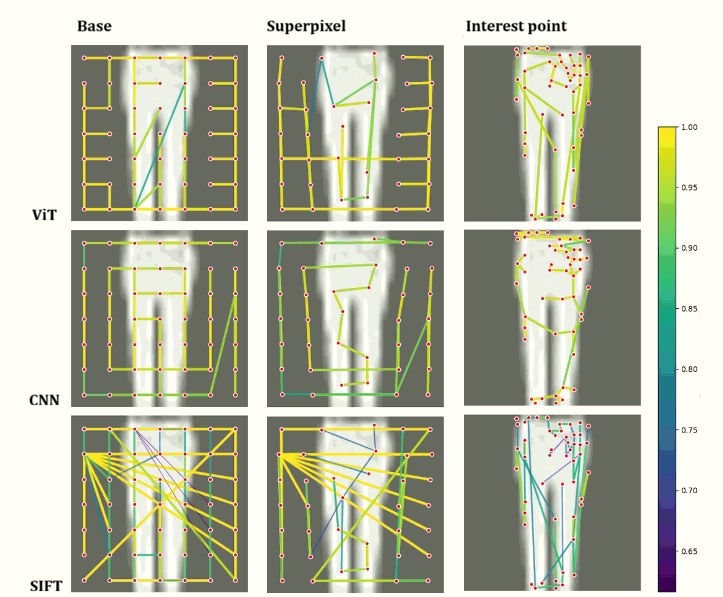}}
\caption{MinCONN Model: final structure of the graph derived from the construction phase.}
\label{fig:ConnectedModel}
\end{figure}

Table \ref{tab:Fashion MNIST Results} presents the top-performing graph model for each node method vs feature type. The overall best result across all experiments is found for interest points node extractor, with ViT features and the MinCONN filter reaching a test accuracy of 0.8681 and an F1 score of 0.8672. For CNN features, the MinCONN filter is again the top performer (accuracy 0.8511, F1 0.8498), while for SIFT features, the optimal filter type deviates, with the LDM filter yielding the best result (accuracy 0.7659, F1 0.7611). The best performance is again obtained with ViT features and the MinCONN filter, achieving a test accuracy of 0.8572 and an F1 score of 0.8581, for superpixel node extractor. For other feature extractors, MinCONN filter is optimal for CNN features (accuracy 0.8364, F1 0.8318), and MinCONN graph remain the best for SIFT features (accuracy 0.7929, F1 0.7912). The highest test accuracy (0.8559) is achieved with ViT features and the MinCONN filter type also for grid extracted nodes. This configuration also yields a strong F1 score of 0.8558. For this node extractor, the optimal filter type varies by feature: LSP works best for CNN features (accuracy 0.8537, F1 0.8529), while ``base'' is optimal for SIFT features (accuracy 0.8284, F1 0.8247). 

The results consistently show that ViT-based features produce the highest-performing graphs, yielding a 1.4\% gain over CNN features and about 4.0\% point gain over SIFT. This performance advantage is partly attributable to their higher dimensionality (768-d vs. 256-d for CNN and 130-d for SIFT), which encodes richer semantic information. In fact, disentangling whether the gains stem from the graph topology design or from stronger node representations is not straightforward. However, a small gap in accuracy is observed despite the descriptor being significantly more compact. An analysis of the optimal configurations reveals that 6 out of 9 best-performing models utilize the MinCONN filter, demonstrating that effective graph-based classification does not require complex or densely connected structures. Fig. \ref{fig:ConnectedModel} illustrates the resulting graph structure. Furthermore, an additional 2 out of 9 optimal configurations employ alternative reduction strategies (LDM and LSP filters), bringing the total to 8 out of 9 best performers that benefit from some form of graph sparsification compared to the base model. A carefully pruned, sparse topology that preserves only the most significant relationships is sufficient to achieve a very good performance. This finding underscores that graph simplicity and structural quality are more critical than connectivity quantity.

These results demonstrate a clear hierarchy in factors influencing classification performance. While the choice of feature extractor establishes the baseline accuracy, the graph architecture, defined by the node extraction method and the edge filter type, plays a critical modulating role, leading to performance variations of up to 4\%.
\section{Conclusion}
This work establishes that graph architecture is not a neutral design choice. It actively modulates the utility of node features. The key insight—that richer features enable sparser, more efficient graphs—provides a principled guideline for resource-constrained deployment. Future work that bridges learned graph construction, efficiency-aware optimization, and theoretical analysis will be essential to translate these findings into robust, scalable graph-based vision systems.
The experimental work has been applied on a limited dataset thus requires a more complex resolution and colour scale setting. Moreover, a limitation of the current study is that we did not incorporate architectural measures for analyzing the neural network's structure. Future work should explore metrics from network analysis to gain deeper insights into how graph topology influences information flow and learning dynamics within the GNN. Finally, a comparison of our results with top performing models was not carried out, as our study is to demonstrate the effectiveness/efficiency of a simpler network architecture.
%
% ---- Bibliography ----
%
% BibTeX users should specify bibliography style 'splncs04'.
% References will then be sorted and formatted in the correct style.
%
\bibliographystyle{splncs04}
\bibliography{main}

\end{document}